\documentclass[10pt,twocolumn,letterpaper]{article}

\usepackage{cvpr}

\usepackage{multirow}
\usepackage{caption}

\definecolor{cvprblue}{rgb}{0.21,0.49,0.74}
\usepackage[pagebackref,breaklinks,colorlinks,allcolors=cvprblue]{hyperref}

\def\paperID{60}
\def\confName{CVPR}
\def\confYear{2026}

\title{NTIRE 2026 The 3rd Restore Any Image Model (RAIM) Challenge: Multi-Exposure Image Fusion in Dynamic Scenes (Track 2)}

\author{
Lishen Qu \hspace{1.7em} Yao Liu \hspace{1.7em} Jie Liang \hspace{1.7em} Hui Zeng \hspace{1.7em} Wen Dai \and
Guanyi Qin \hspace{1.7em} Ya-nan Guan \hspace{1.7em} Shihao Zhou \hspace{1.7em} Jufeng Yang \hspace{1.7em} Lei Zhang \and
Radu Timofte \hspace{1.7em} Xiyuan Yuan \hspace{1.7em} Wanjie Sun \hspace{1.7em} Shihang Li \hspace{1.7em} Bo Zhang \and
Bin Chen \hspace{1.7em} Jiannan Lin \hspace{1.7em} Yuxu Chen \hspace{1.7em} Qinquan Gao \hspace{1.7em} Tong Tong \and
Song Gao \hspace{1.7em} Jiacong Tang \hspace{1.7em} Tao Hu \hspace{1.7em} Xiaowen Ma \hspace{1.7em} Qingsen Yan \and
Sunhan Xu \hspace{1.7em} Juan Wang \hspace{1.7em} Xinyu Sun \hspace{1.7em} Lei Qi \hspace{1.7em} He Xu \and
Jiachen Tu \hspace{1.7em} Guoyi Xu \hspace{1.7em} Yaoxin Jiang \hspace{1.7em} Jiajia Liu \hspace{1.7em} Yaokun Shi
}

\begin{document}
\maketitle

\begin{abstract}
This paper presents NTIRE 2026, the 3rd Restore Any Image Model (RAIM) challenge on multi-exposure image fusion in dynamic scenes. 
We introduce a benchmark that targets a practical yet difficult HDR imaging setting, where exposure bracketing must be fused under scene motion, illumination variation, and handheld camera jitter. 
The challenge data contains 100 training sequences with 7 exposure levels and 100 test sequences with 5 exposure levels, reflecting real-world scenarios that frequently cause misalignment and ghosting artefacts. 
We evaluate submissions with a leaderboard score derived from PSNR, SSIM, and LPIPS, while also considering perceptual quality, efficiency, and reproducibility during the final review. 
This track attracted 114 participating teams and received 987 submissions. The winning methods significantly improved the ability to remove artifacts from multi-exposure fusion and recover fine details.
The dataset and the code of each team can be found at the repository: \url{https://github.com/qulishen/RAIM-HDR}.
%
\end{abstract}

\section{Introduction}
\label{sec:introduction}
Image restoration and enhancement aim to recover visually pleasing images from degraded inputs and have become core components of modern computational photography and mobile imaging systems~\cite{kalantari2017deep,shu2024towards_realhdrv,jiang2023meflut}. Among these problems, multi-exposure image fusion (MEF) and high dynamic range (HDR) reconstruction are widely used to extend the dynamic range of commodity cameras. By combining frames captured at different exposure levels, these methods recover details in both highlights and shadows and produce results beyond the range of a single shot~\cite{chen2025ultrafusion,li2025afunet,kong2024safnet}.

Real-world photography is often far more challenging than the static settings used in many academic benchmarks. Moving objects, camera shake, local lighting changes, and exposure differences can introduce severe misalignment, ghosting, color distortion, and structural artifacts. As a result, dynamic-scene MEF remains considerably harder than fusion in controlled static environments~\cite{kalantari2017deep,sice_cai2018learning,liu2022ghost-free_hdrtransformer}. Practical deployment further requires efficient and stable processing on high-resolution inputs.

MEF and HDR reconstruction address this problem from different domains. MEF usually operates in the non-linear sRGB domain and produces display-ready results, while HDR reconstruction works in the linear domain and preserves radiometric information for later image signal processing~\cite{chen2025ultrafusion}. Recent learning-based methods such as MEFLUT~\cite{jiang2023meflut}, RetinexMEF~\cite{bai2025retinex}, and UltraFusion~\cite{chen2025ultrafusion} have improved detail preservation and robustness. HDR methods~\cite{kong2024safnet,li2025afunet,tel2023alignment} have also reduced motion artifacts with optical flow, CNNs, and Transformers, although many solutions still rely on heavy alignment components.

A clear gap nevertheless remains between academic research and practical deployment. Real camera pipelines, especially on smartphones, face diverse degradations, limited memory, and strict latency requirements. The Restore Any Image Model (RAIM) challenge series was launched to narrow this gap by focusing on real-world image restoration under practical efficiency constraints. Previous RAIM editions targeted restoration tasks in RGB and RAW domains and attracted broad academic and industrial participation.

In NTIRE 2026, we extend the RAIM initiative to dynamic-scene multi-exposure image fusion, which is highly relevant to mobile photography. We organize Track 2 around a new benchmark that emphasizes motion robustness, detail fidelity, artifact suppression, and model efficiency. The challenge is co-hosted by the Y-Lab of OPPO Research Institute, the Computer Vision Lab of Nankai University, and the Visual Computing Lab of The Hong Kong Polytechnic University.

This report summarizes the challenge motivation, dataset design, evaluation protocol, submission rules, participating methods, and final results. We aim to establish a reproducible benchmark for dynamic-scene MEF and to encourage efficient and generalizable solutions for real-world mobile HDR imaging.

This challenge is one of the challenges associated with the NTIRE 2026 Workshop~\footnote{  
  \url{https://www.cvlai.net/ntire/2026/} 
  
  \ \ Lishen Qu, Yao Liu, Jie Liang, Hui Zeng, Wen Dai, Guanyi Qin, Ya-nan Guan, Shihao Zhou, Jufeng Yang, Lei Zhang, and Radu Timofte are the organizers of the NTIRE 2026 challenge (RAIM Track 2), and other authors are the participants. 
  
  \ \ The Appendix lists the authors’ teams and affiliations.  
} on:
deepfake detection~\cite{ntire26deepfake}, 
high-resolution depth~\cite{ntire26hrdepth},
multi-exposure image fusion~\cite{ntire26raim_fusion}, 
AI flash portrait~\cite{ntire26raim_portrait}, 
professional image quality assessment~\cite{ntire26raim_piqa},
light field super-resolution~\cite{ntire26lightsr},
3D content super-resolution~\cite{ntire263dsr},
bitstream-corrupted video restoration~\cite{ntire26videores},
X-AIGC quality assessment~\cite{ntire26XAIGCqa},
shadow removal~\cite{ntire26shadow},
ambient lighting normalization~\cite{ntire26lightnorm},
controllable Bokeh rendering~\cite{ntire26bokeh},
rip current detection and segmentation~\cite{ntire26ripdetseg},
low light image enhancement~\cite{ntire26llie},
high FPS video frame interpolation~\cite{ntire26highfps},
Night-time dehazing~\cite{ntire26nthaze,ntire26nthaze_rep},
learned ISP with unpaired data~\cite{ntire26isp},
short-form UGC video restoration~\cite{ntire26ugcvideo},
raindrop removal for dual-focused images~\cite{ntire26dual_focus},
image super-resolution (x4)~\cite{ntire26srx4},
photography retouching transfer~\cite{ntire26retouching},
mobile real-word super-resolution~\cite{ntire26rwsr},
remote sensing infrared super-resolution~\cite{ntire26rsirsr},
AI-Generated image detection~\cite{ntire26aigendet},
cross-domain few-shot object detection~\cite{ntire26cdfsod},
financial receipt restoration and reasoning~\cite{ntire26finrec},
real-world face restoration~\cite{ntire26faceres},
reflection removal~\cite{ntire26reflection},
anomaly detection of face enhancement~\cite{ntire26anomalydet},
video saliency prediction~\cite{ntire26videosal},
efficient super-resolution~\cite{ntire26effsr},
3d restoration and reconstruction in adverse conditions~\cite{ntire26realx3d},
image denoising~\cite{ntire26denoising},
blind computational aberration correction~\cite{ntire26aberration},
event-based image deblurring~\cite{ntire26eventblurr},
efficient burst HDR and restoration~\cite{ntire26bursthdr},
low-light enhancement: `twilight cowboy'~\cite{ntire26twilight},
and efficient low light image enhancement~\cite{ntire26effllie}.

\section{Challenge Data}
\label{sec:data}


\begin{figure*}[t]
    \centering
    \includegraphics[width=\linewidth]{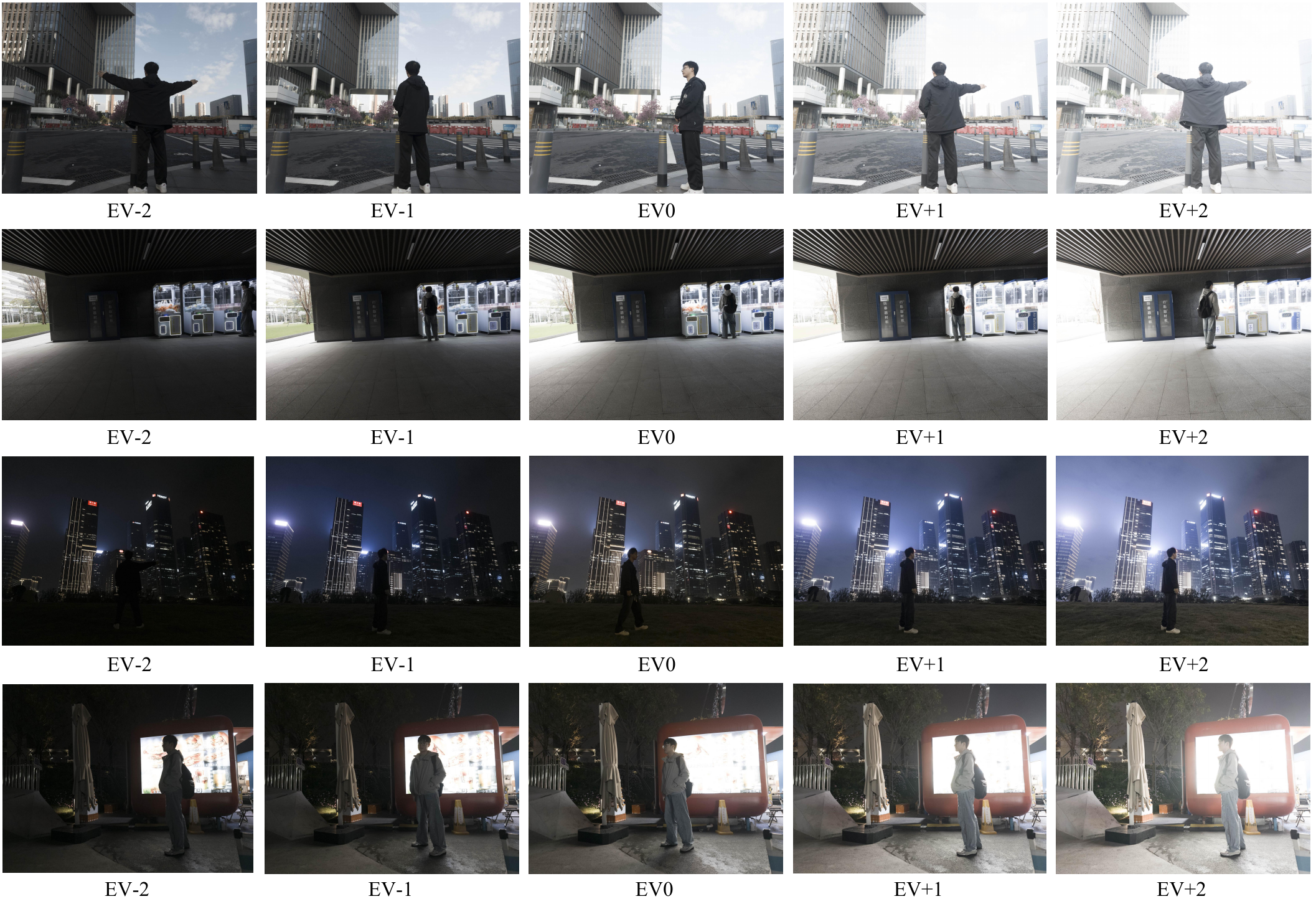}
    \caption{Representative image sequences in RAIM MEF in dynamic scenes. The data includes various scenes and times, with a particular focus on scenes with high dynamic range.}
    \label{fig:data_test}
\end{figure*}

Representative training and test samples are shown in~\cref{fig:data_test}. The Track 2 dataset is designed to reflect dynamic multi-exposure capture in realistic photographic conditions. Each sequence contains bracketed exposures of the same scene, while the scene may include moving objects, local appearance changes, and camera shake between frames. The benchmark evaluates not only fusion quality but also robustness to misalignment and ghosting artifacts.

\noindent\textbf{Training set.} We provide 100 multi-exposure sequences for model development. Each training sequence contains 7 exposure levels, enabling participants to study both the effect of exposure diversity and the robustness of their fusion pipeline under dynamic motion. The data is paired with reference fusion images for supervised training and validation of perceptual quality.

\noindent\textbf{Test set.} The test benchmark contains 100 sequences, each with 5 exposure levels. Compared with the training set, the reduced number of frames makes the fusion problem more constrained and encourages methods that remain stable when less exposure information is available.

\noindent\textbf{Benchmark characteristics.} The dataset includes several sources of difficulty that are central to real-world HDR fusion
\begin{itemize}
    \item object motion across exposure levels
    \item illumination variation between frames
    \item handheld camera jitter and global misalignment
    \item local saturation and underexposure in different regions
\end{itemize}

These properties make the benchmark suitable for evaluating practical multi-exposure fusion models that must balance detail recovery, motion handling, and visual naturalness in real-world scenarios.

\section{Evaluation Protocol}
\label{sec:evaluation}

We evaluate the submitted methods using both objective image-quality metrics and organizer-side reproducibility checks. The public leaderboard score follows the official quantitative protocol released for the challenge. Specifically, we compute PSNR, SSIM, and LPIPS with respect to the private reference fusion images, and combine them into a single ranking score as
\begin{equation}
\mathrm{Score} = 30 \cdot \frac{\mathrm{PSNR}}{50}
+ 22.5 \cdot \frac{\mathrm{SSIM} - 0.5}{0.5}
+ 30 \cdot \left(1 - \frac{\mathrm{LPIPS}}{0.4}\right).
\end{equation}

Beyond the main ranking score, the challenge also records DISTS, NIQE, subjective quality following~\cite{guan1,guan2}, and efficiency measures such as runtime, parameter count, and FLOPs.
We include these dimensions because a practically useful fusion method should not only achieve a high score against the reference images, but also remain efficient and visually reliable in real-world deployment.

\noindent\textbf{Phase-2 submission format.} Participants are required to submit a single \texttt{.zip} archive containing exactly 100 JPEG files named from \texttt{001.jpg} to \texttt{100.jpg}.
The archive must also include a \texttt{readme.txt} file that reports runtime per image, whether the method uses CPU or GPU at test time, whether extra training data is used, and any additional notes about the submitted solution.

\noindent\textbf{Final-phase evaluation.} For the final phase, contestants must additionally submit inference code and pre-trained weights in a compressed package that includes a runnable \texttt{test.sh} script. The final ranking is determined only among participants whose code is submitted on time, passes the organizer review, and can be reproduced locally within a reasonable range of the leaderboard performance. The competition offers awards including one first-class prize of \textbf{US\$1000}, two second-class prizes of \textbf{US\$500} each, and three third-class prizes of \textbf{US\$200} each.



\section{Submitted Methods}
\label{sec:methods}

\subsection{WHU-VIP}

\begin{figure}[t]
    \centering
    \includegraphics[width=\linewidth]{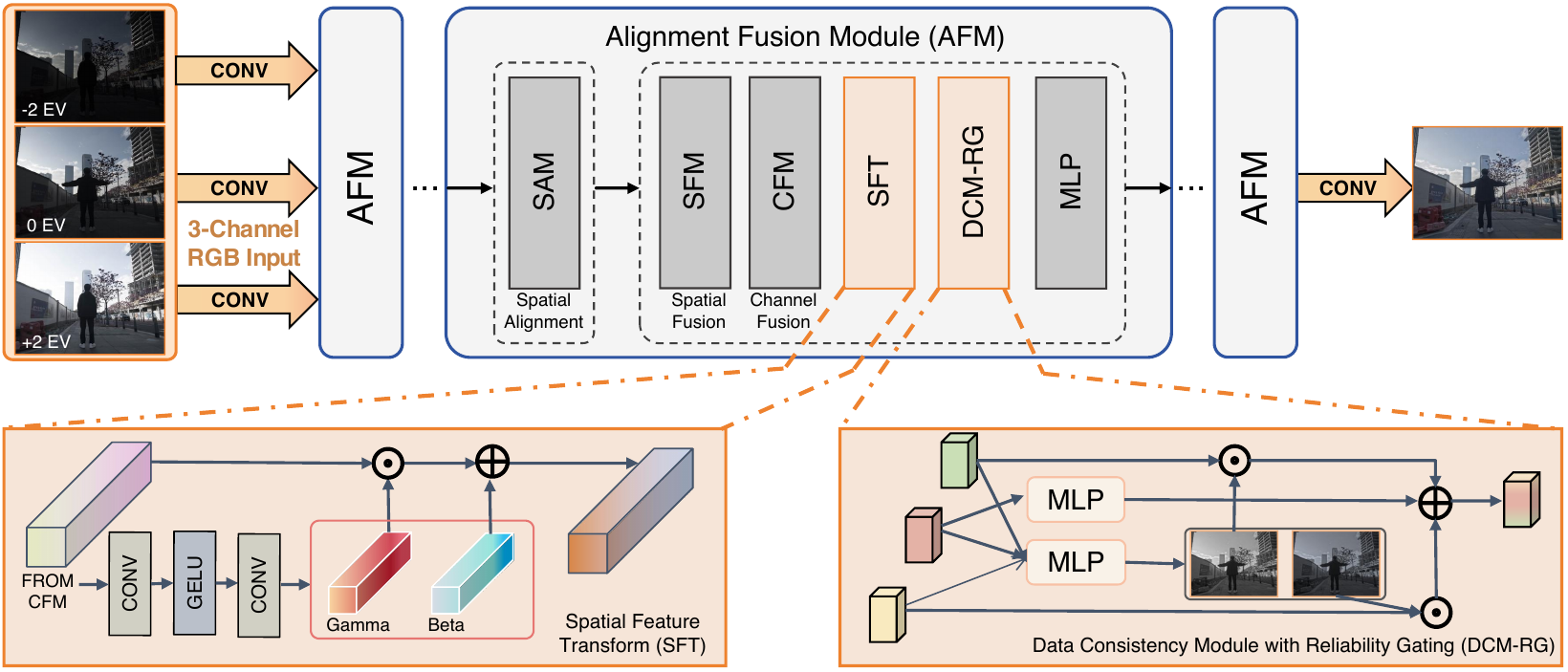}
    \caption{Overview of the WHU-VIP method with the AFUNet backbone, the SFT module, and the DCM-RG module.}
    \label{fig:whu_vip_modules}
\end{figure}

\noindent\textbf{Overview and input formulation.}
The WHU-VIP team builds its method on AFUNet~\cite{li2025afunet}, which uses the Swin Transformer to model long-range spatial dependencies. Although the dataset provides seven multi-exposure images for training and five for testing, the team uses only three frames as input, namely the reference frame at $0$ EV and two auxiliary frames at $\pm2$ EV. This design improves efficiency while reducing motion-induced misalignment artifacts in dynamic scenes.

The overall pipeline and the detailed designs of the Spatial Feature Transform (SFT) module~\cite{wang2018recovering} and the Data Consistency Module with Reliability Gating (DCM-RG) are shown in~\cref{fig:whu_vip_modules}.

\noindent\textbf{Model modifications.}
To better adapt AFUNet for multi-exposure fusion in dynamic scenes, the WHU-VIP team introduces three modifications.
\begin{itemize}
    \item \textit{Network I/O adaptations.} Because the image signal processing pipeline of the dataset is unknown, the team removes the original 6-channel gamma-approximated input design and directly processes standard 3-channel RGB images. The final Sigmoid activation is also removed because the target fused image lies in the same non-linear image space as the inputs.
    
    \item \textit{Spatial Feature Transform.} An SFT module is inserted after the Channel Fusion Module to perform spatially adaptive feature modulation. This design improves fusion in regions affected by motion occlusions or extreme exposures by adjusting feature responses according to local illumination conditions.
    
    \item \textit{Reliability gating.} To suppress artifacts caused by misalignment or unreliable exposures, the team introduces DCM-RG. This module replaces fixed scalar fusion weights with a reliability gating mechanism and predicts pixel-wise spatial gates $g_1, g_2 \in [0,1]$ to suppress unreliable regions and prioritize robust features during fusion.
\end{itemize}

\noindent\textbf{Data curation and preprocessing.}
To stabilize training, the WHU-VIP team first evaluates all scenes with a pilot model and removes the 10 lowest-scoring scenes that contain severe degradations. The remaining 90 scenes are cropped offline into overlapping $512 \times 512$ patches and then randomly cropped to $256 \times 256$ patches during training.

\noindent\textbf{Objective function.}
Inspired by the official evaluation metric, the WHU-VIP team defines a simplified score function.
$$
\mathcal{S} = w_{1} \mathrm{PSNR} + w_{2} \mathrm{SSIM} - w_{3} \mathrm{LPIPS}.
$$
Initially, the weights were set to $(w_1, w_2, w_3) = (0.6, 45.0, 75.0)$ to emphasize structural fidelity and stable convergence. In later stages, they were adjusted to $(0.1, 5.0, 187.5)$ to encourage perceptually realistic high-frequency details. The final optimization objective is defined as $\mathcal{L} = 1 - \mathcal{S}_{\mathrm{norm}}$, where $\mathcal{S}_{\mathrm{norm}}$ denotes the min-max normalized score within $[0,1]$.

\noindent\textbf{Curriculum learning strategy.}
To avoid poor local minima and progressively improve reconstruction quality, the WHU-VIP team uses a four-stage curriculum
\begin{itemize}
\item \textit{Stage 1. Baseline alignment.} The model is trained from scratch with basic augmentations and the initial objective weights to establish reliable feature alignment and fusion.

\item \textit{Stage 2. Dynamic generalization.} Stronger augmentations, including local exposure perturbation, global gamma jitter, spatial misalignment, and channel shuffle, are introduced to simulate challenging dynamic scenes and improve robustness to extreme exposures and illumination variations.

\item \textit{Stage 3. Perceptual enhancement.} The objective weights are adjusted to emphasize LPIPS and encourage visually realistic outputs with improved high-frequency textures.

\item \textit{Stage 4. Quantization-aware training.} A Straight-Through Estimator is used for 8-bit quantization-aware training to mitigate truncation errors when exporting float32 predictions as 8-bit images.
\end{itemize}

An exponential moving average of the model weights is maintained throughout training to stabilize optimization. This staged strategy first establishes structural alignment and then refines perceptual details.

\subsection{SHL}

\begin{figure}[h]
    \centering
    \includegraphics[width=0.95\columnwidth]{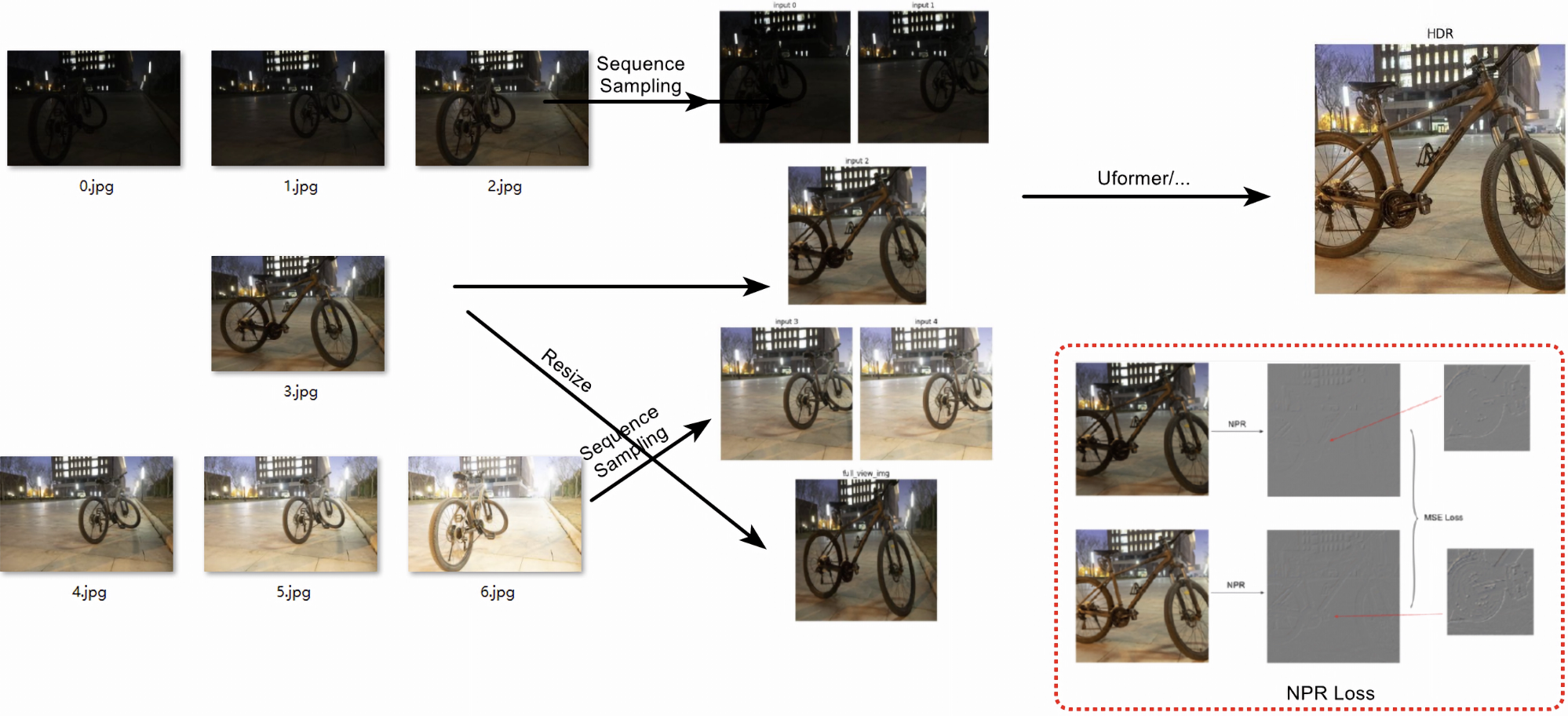}
    \caption{Overview of the SHL team's proposed framework. The method introduces a lightweight global guidance branch that provides context from the downsampled full-resolution image, complemented by a Neighboring Pixel Relationships (NPR) loss to preserve high-frequency local details.}
    \label{fig:SHL_track2}
\end{figure}

The SHL team proposes a lightweight global guidance strategy that achieves performance enhancement of established models with minimal computational overhead. Specifically, the full-resolution image is downsampled to the training scale and concatenated to the model input, while the Neighboring Pixel Relationships (NPR) Loss is introduced to constrain local structural consistency during training.

\noindent\textbf{Method description.}
The SHL team proposes a lightweight global guidance strategy designed to enhance the performance of established restoration models with minimal computational overhead. To address the limited receptive field of patch-based training, they downsample the full-resolution input image to the training scale and concatenate it with the model input. This introduces global structural priors that guide the network during inference. Furthermore, to mitigate the blurring artifacts common in pixel-wise optimization, the team introduces a Neighboring Pixel Relationships (NPR) Loss. This loss function explicitly constrains the consistency of local structures, ensuring that high-frequency details and edge integrity are preserved in the final output.

\noindent\textbf{Training and testing details.}
Experiments were conducted on a single RTX 4090 GPU with 24 GB memory. The input tensor size is $768 \times 768 \times 18$, formed by concatenating five key frames sampled symmetrically from the original 7-frame exposure sequence together with one downsampled intermediate frame as a global reference. The SHL team applies random resizing with a minimum edge of 768 pixels before random cropping and uses horizontal flips, vertical flips, and 90-degree rotations for augmentation. Three backbones, namely Restormer~\cite{zamir2022restormer}, Uformer~\cite{Wang_2022_CVPR}, and MST++~\cite{mst}, are trained from scratch on the organizer-provided dataset. Optimization uses Adam for 60k iterations with a batch size of 1 and a MultiStepLR schedule. The training loss combines Charbonnier, SSIM, gradient, color, perceptual, and NPR terms. During inference, the team uses sliding-window cropping with a 1528-pixel crop size and fuses the predictions of the three models by weighted averaging.

\subsection{nunucccb}
\begin{figure}[h]
    \centering
    \includegraphics[width=1\columnwidth]{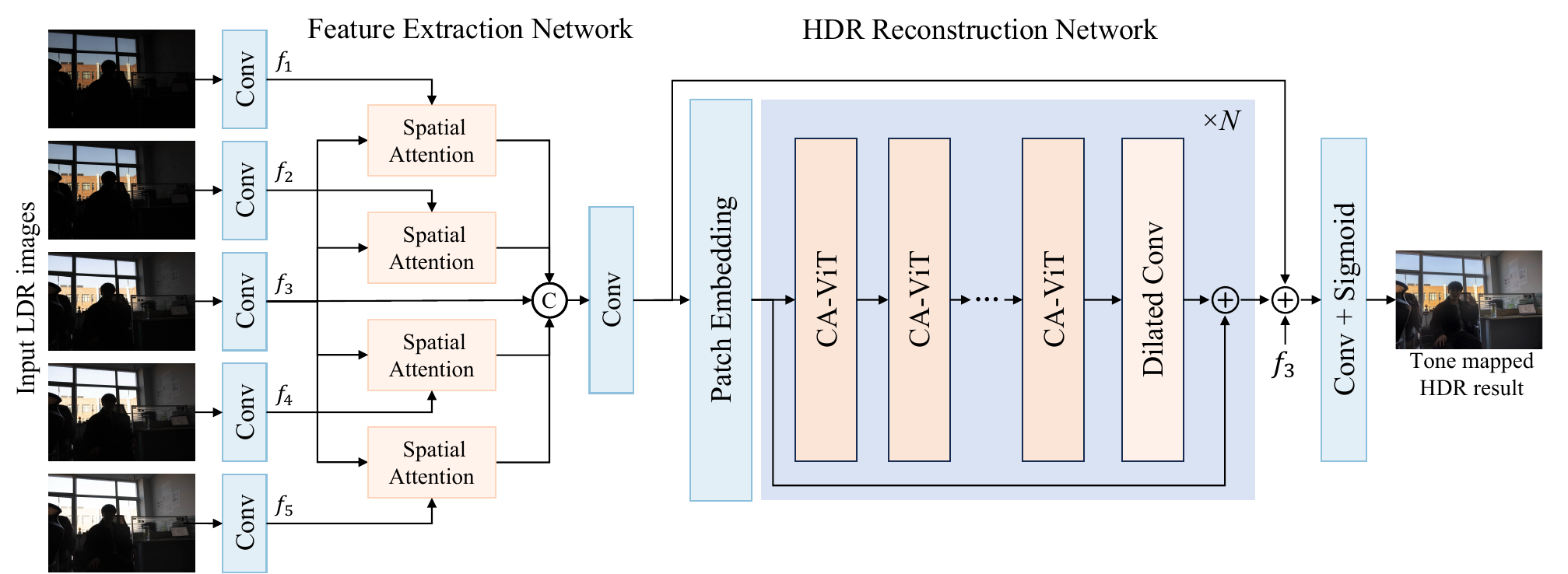}
    \caption{Overview of the nunucccb method.}
    \label{fig1}
\end{figure}

The nunucccb team adopts the HDR-Transformer framework based on the Context-Aware Vision Transformer~\cite{liu2022ghost-free_hdrtransformer}.

\noindent \textbf{Network architecture.} The model contains two main components as shown in~\cref{fig1}, namely a feature extraction network and a reconstruction network. In the feature extraction network, independent convolution layers first extract shallow features ($f_1$ to $f_5$) from the five input images. Taking $f_3$ as the reference feature, spatial attention modules are applied to the non-reference features ($f_1, f_2, f_4, f_5$) to effectively suppress undesired misalignments caused by foreground movements. These aligned features and the reference feature $f_3$ are then concatenated and coarsely fused through a subsequent convolution layer. 

The reconstruction network takes the fused features and processes them through a patch embedding layer, followed by a sequence of $N$ CA-ViT blocks to model complex global dependencies and intensity variations. After the CA-ViT blocks, a dilated convolution layer is employed to further expand the receptive field. Finally, the structural integrity of the reference feature $f_3$ is preserved through residual connections, followed by a final convolution layer to yield the HDR result.

\noindent \textbf{Training details.} During training, the nunucccb team randomly crops $256 \times 256$ patches from the training data and augments them with horizontal and vertical flips. All experiments are conducted on a single NVIDIA L40 GPU for 150,000 iterations. The network is optimized with Adam using $\beta_1=0.9$ and $\beta_2=0.999$, and the initial learning rate of $2 \times 10^{-4}$ is gradually decayed with a CosineAnnealing scheduler.

The overall objective function combines the pixel-level $\mathcal{L}_{1}$ reconstruction loss and the VGG19-based perceptual loss.

\noindent \textbf{Additional information.} The team uses only the official dataset provided by the challenge organizers and does not rely on external data.

\subsection{untrafusion}

\begin{figure}[t]
    \centering
    \includegraphics[width=\linewidth]{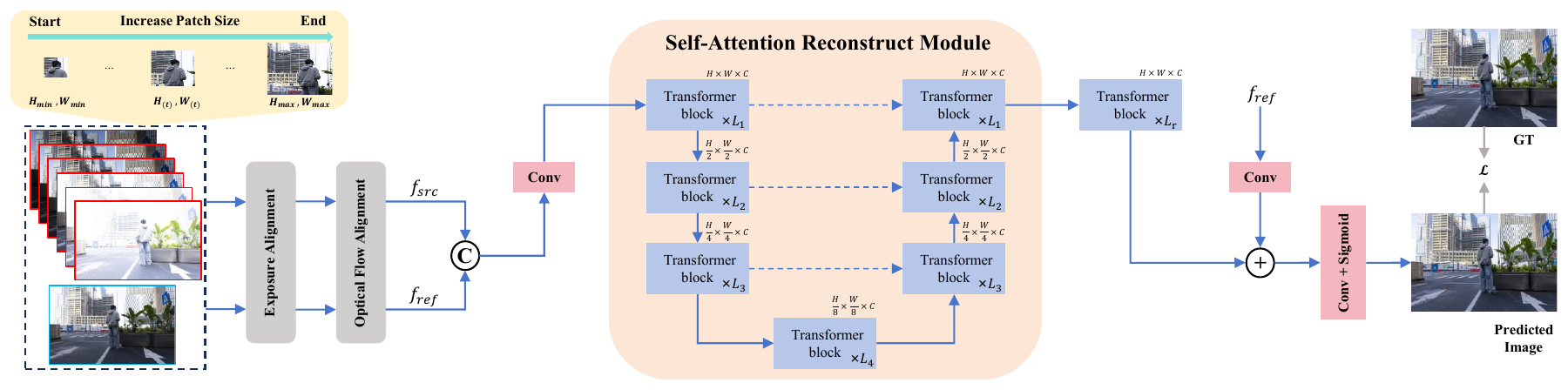}
    \caption{Overview of the untrafusion pipeline.}
    \label{fig:DeepTrans}
\end{figure}

\noindent\textbf{General method description.}
The untrafusion team proposes an end-to-end reconstruction framework that explicitly models physical exposure priors and uses a scale-evolving training strategy to address noise sensitivity and structural artifacts in the non-linear JPG domain. The overall architecture is shown in~\cref{fig:DeepTrans}.

The baseline follows a multi-frame feature fusion paradigm that can handle an arbitrary number of input frames and therefore accommodates the difference between the 7-frame training setting and the 5-frame test setting. The framework contains a motion compensation stage followed by a hierarchical reconstruction backbone. For inter-frame registration, the team uses a coarse-to-fine optical flow alignment module~\cite{ranjan2017optical}. The reconstruction module uses a symmetric encoder-decoder structure based on Restormer~\cite{zamir2022restormer}, where three downsampling operations expand the receptive field and three matching upsampling stages restore spatial resolution. Refinement transformer blocks are further appended after the UNet core to improve high-frequency reconstruction.

To mitigate severe misalignment caused by large pixel-value differences across exposures, the team introduces a luminance-prior guided exposure alignment module. The dynamic range of each non-reference frame is linearly mapped to the reference luminance level using statistically derived luminance ratios. To balance global structure and local detail under limited computation, the team also adopts a curriculum-based progressive tiling strategy~\cite{fischer2025progressive}. The model is first warmed up on small patches and is then trained with progressively larger patches.

\noindent\textbf{Implementation details.}
During training, the patch size is increased from $256^2$ to $512^2$ and finally $768^2$ to suppress artifacts. Random flipping and rotation are also used for augmentation~\cite{shorten2019survey}.
The framework was implemented in PyTorch and optimized via the Adam algorithm ($\beta_1=0.9, \beta_2=0.999$). The learning rate followed a Cosine Annealing schedule, starting at $2\times10^{-4}$ and decaying to $2\times10^{-6}$ over 120 epochs.
The training was partitioned into three phases with increasing patch sizes, lasting for 20, 30, and 70 epochs, respectively.
For evaluation, the team uses TLC~\cite{chu2022improving} to ensure seamless full-image inference without memory constraints.

\subsection{\texorpdfstring{$I^2$ Group \& Transsion}{I2 Group \& Transsion}}
\begin{table*}[!t]\footnotesize
    \centering
    \caption{Verified final results for RAIM Track 2. The ranking is computed from validated scores on Test Stage 1 and Test Stage 2.}
    \label{tab:final_results}
    \begin{tabular}{llccccccccccc}
    \toprule
    \multirow{2}{*}{Team Name} & \multicolumn{4}{c}{Test Stage 1} &  & \multicolumn{4}{c}{Test Stage 2} &  & \multirow{2}{*}{Final Score $\uparrow$} \\ \cline{2-5} \cline{7-10}
     & PSNR $\uparrow$& SSIM $\uparrow$& LPIPS $\downarrow$& Score $\uparrow$&  & PSNR $\uparrow$& SSIM $\uparrow$& LPIPS $\downarrow$& Score $\uparrow$&  &  \\ \cline{1-12} 
    WHU-VIP & 27.059 & 0.915 & 0.089 & 58.249 &  & 27.556 & 0.923 & 0.080 & 59.529 &  & 58.889 \\
    SHL & 27.653 & 0.920 & 0.112 & 57.110 &  & 28.825 & 0.928 & 0.101 & 58.958 &  & 58.034 \\
    nunucccb & 26.839 & 0.913 & 0.111 & 56.395 &  & 27.011 & 0.918 & 0.099 & 57.605 &  & 57.000 \\
    untrafusion & 26.847 & 0.908 & 0.095 & 57.333 &  & 26.002 & 0.907 & 0.122 & 54.733 &  & 56.033 \\
    I\textasciicircum{}2 Group \& Transsion & 26.221 & 0.904 & 0.121 & 54.853 &  & 26.137 & 0.909 & 0.113 & 55.619 &  & 55.236 \\
    miketjc & 26.002 & 0.907 & 0.122 & 54.733 &  & 26.312 & 0.909 & 0.118 & 55.343 &  & 55.038 \\
    \bottomrule
    \end{tabular}
\end{table*}
\begin{figure}[h]
    \centering
    \includegraphics[width=0.95\columnwidth]{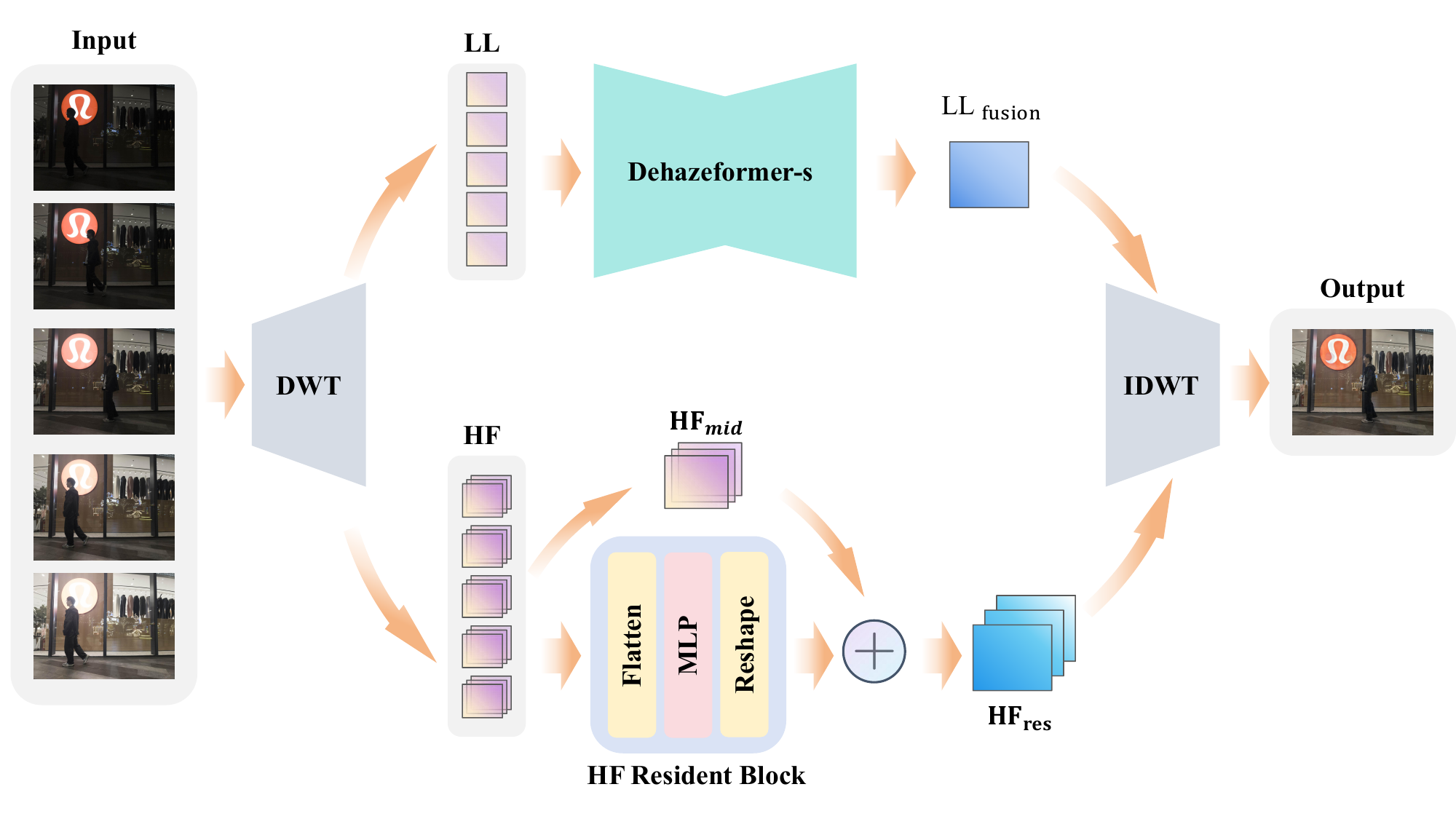}
    \vspace{-3mm}
    \caption{Overview of the $I^2$ Group \& Transsion method.}
    \label{fig:team_name}
    
\end{figure}

The $I^2$ Group \& Transsion team proposes a wavelet-based multi-exposure fusion network that processes low-frequency and high-frequency information in separate branches. Given a sequence of five multi-exposure images, the team first applies the discrete wavelet transform~\cite{liu2018multi} with Haar wavelets to decompose each input into low-frequency and high-frequency components. This decomposition separates structural information from fine details.

For the low-frequency branch, the team uses \textbf{DehazeFormer-s}~\cite{song2023vision} as the backbone. This architecture preserves intensity information and captures long-range dependencies through adaptive normalization and window-based attention. The input layer is modified to accept 15 channels formed by concatenating the five exposures.

In the high-frequency branch, directly fusing information from all exposures tends to introduce ghosting, while using only the middle exposure loses details. To address this trade-off, the team designs a lightweight \textbf{HF Residual Fusion} module. The high-frequency component of the middle exposure is used as the reference, and the high-frequency components of all frames are used to learn a residual map. This residual map is added to the reference features to refine high-frequency details before the fused low-frequency and high-frequency features are reconstructed through the inverse DWT.

\subsection{NTR}

\begin{figure}[h]
\centering
\includegraphics[width=\linewidth]{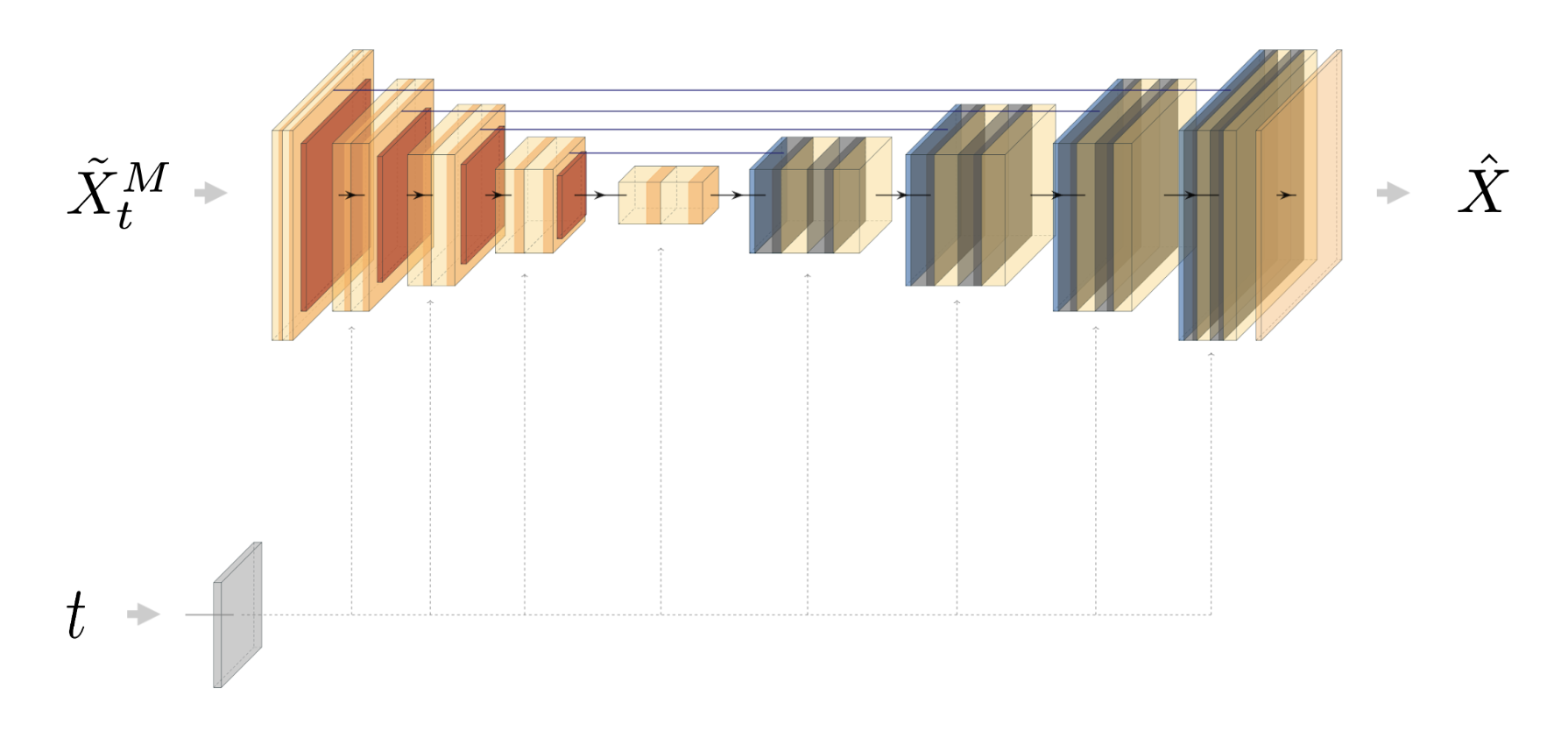}
\vspace{-3mm}
\caption{TimeDiffiT architecture~\cite{tu2025score}. The doubly-corrupted input
$\tilde{X}_t^M$ and noise level $\sigma_t$ enter the time-conditioned U-Net
(encoder: orange; decoder: blue), producing the restored output $\hat{X}$.}
\label{fig:arch}
\vspace{-3mm}
\end{figure}

The NTR team proposes a metric-aligned multi-exposure fusion framework built on a time-conditioned encoder-decoder, as shown in~\cref{fig:arch}. The method takes five LDR exposures as input, concatenates them into a 15-channel tensor, and predicts the fused RGB image in a single forward pass. The core design emphasizes consistency with the official challenge metric so that training better matches the final ranking criterion.

\noindent\textbf{Method description.}
The NTR team initializes the backbone from a pretrained masked diffusion autoencoder and then adapts it to the MEF task via supervised fine-tuning. The model uses a time-conditioned U-Net architecture with a 15-channel input stem and a skip projection from the input stack to the output image. This design allows the network to learn a global exposure blend while correcting local motion artifacts and ghosting.

\noindent\textbf{Training strategy.}
Training is performed in two stages. In the first stage, the pretrained model is adapted to the MEF domain with supervised denoising fine-tuning. In the second stage, the NTR team optimizes a composite loss that combines $\ell_1$, SSIM, and LPIPS terms so that the objective better reflects the official score. The team further tunes the perceptual weight through continuation training and reports consistent gains from this metric-aligned strategy.

\noindent\textbf{Training and inference details.}
The NTR team trains on 90 scenes and selects five evenly spaced exposures to match the test setting. Random $256 \times 256$ crops, horizontal flips, vertical flips, and transpose augmentation are used during training. At inference time, full-resolution images are processed with tiled prediction, and overlapping outputs are averaged. The final submission also uses an 8-way geometric self-ensemble~\cite{timofte2016seven} to improve robustness and final score.

\begin{figure*}[t]
    \centering
    \includegraphics[width=0.95\linewidth]{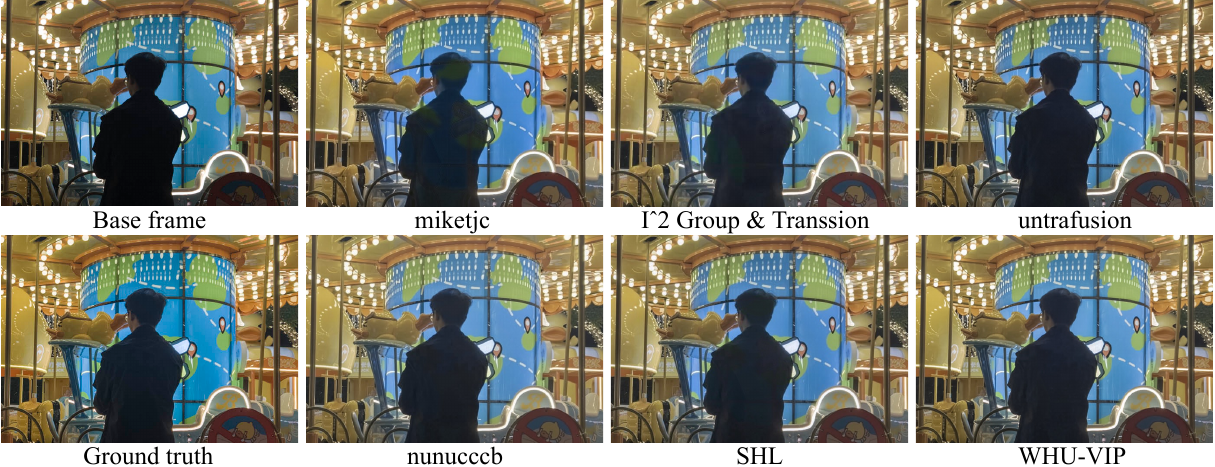}
    \caption{Visual comparison on representative examples from Test Stage 1. The compared methods show clear differences in structural consistency, detail preservation, and ghosting suppression.}
    \label{fig:com1}
\end{figure*}

\begin{figure*}[t]
    \centering
    \includegraphics[width=0.95\linewidth]{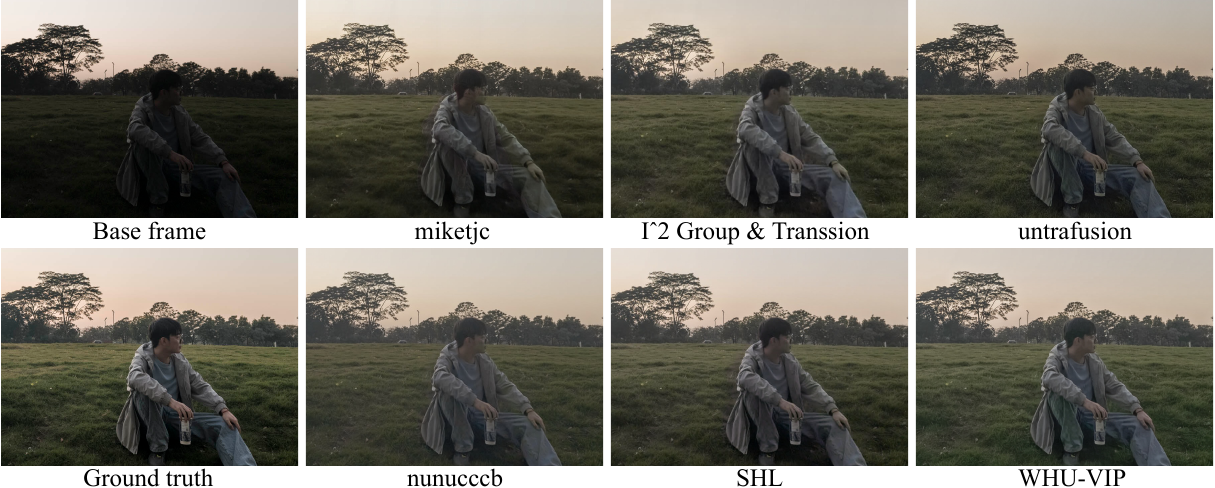}
    \caption{Visual comparison on representative examples from Test Stage 2. This stage further highlights differences in motion handling, exposure robustness, and artifact suppression.}
    \label{fig:com2}
\end{figure*}

\section{Results Overview}

\label{sec:results}
\noindent\textbf{Quantitative comparison.} As shown in~\cref{tab:final_results}, WHU-VIP ranks first overall with a final score of 58.889 and achieves the best balance across the two evaluation stages. Although SHL records the highest PSNR and SSIM on both stages, WHU-VIP obtains consistently lower LPIPS values and the highest stage scores, which leads to the best overall ranking under the official metric. The nunucccb team ranks third and also maintains stable performance across both stages, indicating competitive generalization under the challenge protocol.

\noindent\textbf{Cross-stage analysis.} The two-stage evaluation reveals that leaderboard positions are influenced not only by peak fidelity on a single subset but also by robustness across different test conditions. In particular, the top three teams remain relatively stable from Test Stage 1 to Test Stage 2, while several lower-ranked submissions show a more visible performance drop on the second stage. This trend suggests that handling exposure variation, scene motion, and local misalignment in a consistent manner remains a key difficulty for multi-exposure fusion in dynamic scenes.

\noindent\textbf{Visual comparison.} As shown in~\cref{fig:com1,fig:com2}, the representative qualitative results for Test Stage 1 and Test Stage 2 are broadly consistent with the quantitative ranking. Stronger methods usually produce cleaner structures, fewer ghosting artifacts, and better local detail preservation in challenging dynamic regions. These examples also show that high-performing solutions must preserve perceptual naturalness and structural fidelity at the same time.

\section{Teams and Affiliations}
\label{sec:teams}

\flushleft

\noindent\textit{\textbf{Organizers}}

\noindent Lishen Qu$^{1,2}$ (qulishen@mail.nankai.edu.cn)

\noindent Yao Liu$^{1,2}$ (liuyao@mail.nankai.edu.cn)

\noindent Jie Liang$^1$ (liang27jie@163.com)

\noindent Hui Zeng$^{1}$ (cshzeng@gmail.com)

\noindent Wen Dai$^{1}$ (daiwen@oppo.com)

\noindent Guanyi Qin$^{1,5}$ (guanyi.qin@u.nus.edu)

\noindent Ya-nan Guan$^{1,2}$ (guanyanan@mail.nankai.edu.cn)

\noindent Shihao Zhou$^{1,2}$ (zhoushihao96@mail.nankai.edu.cn)

\noindent Prof. Jufeng Yang$^{2}$ (yangjufeng@nankai.edu.cn)

\noindent Prof. Lei Zhang$^{1,3}$ (cslzhang@comp.polyu.edu.hk)

\noindent Prof. Radu Timofte$^4$ (radu.timofte@uni-wuerzburg.de)

\noindent\textit{\textbf{Affiliations}}

\noindent $^1$ OPPO Research Institute

\noindent $^2$ Nankai University

\noindent $^3$ The Hong Kong Polytechnic University 

\noindent $^4$ Computer Vision Lab, University of W\"urzburg, Germany

\noindent $^5$ National University of Singapore

~\\

\noindent \textit{\textbf{Team name:}} WHU-VIP\\
\textit{\textbf{Members:}}
Xiyuan Yuan$^1$ (yuanxiyuan@whu.edu.cn), Wanjie Sun$^{1}$ (sunwanjie@whu.edu.cn)\\
\textit{\textbf{Affiliations}}\\
$^1$ School of Remote Sensing and Information Engineering, Wuhan University, China\\

~\\

\noindent \textit{\textbf{Team name:}} SHL\\
\textit{\textbf{Members:}}
Shihang Li$^1$ (lishihang@gml.ac.cn), Bo Zhang$^1$\\
\textit{\textbf{Affiliations}}\\
$^1$ Guangdong Laboratory of Artificial Intelligence and Digital Economy (SZ), Shenzhen, China\\

~\\

\noindent \textit{\textbf{Team name:}} nunucccb\\
\textit{\textbf{Members:}}
Bin Chen$^1$(nunucccb@gmail.com) 
Jiannan Lin$^1$, Yuxu Chen$^1$, Qinquan Gao$^{1,2}$, Tong Tong$^{1,2}$\\
\textit{\textbf{Affiliations}}\\
$^1$ Fuzhou University\\
$^2$ Imperial Vision Technology \\

~\\

\noindent \textit{\textbf{Team name:}} untrafusion\\
\textit{\textbf{Members:}}
Song Gao$^1$ (gaosong\_cs@mail.nwpu.edu.cn), Jiacong Tang$^1$, Tao Hu$^{1}$, Xiaowen Ma$^{1}$, Qingsen Yan$^{1,2}$\\
\textit{\textbf{Affiliations}}\\
$^1$ School of Computer Science, Northwestern Polytechnical University, China\\
$^2$ Shenzhen Research Institute of Northwestern Polytechnical University

~\\

\noindent \textit{\textbf{Team name:}} $I^2$ Group \& Transsion\\
\textit{\textbf{Members:}}
\noindent Sunhan Xu$^{1,2}$ (25110128@bjtu.edu.cn) \\

\noindent Juan Wang$^2$ (jun\_wang@ia.ac.cn) \\

\noindent Xinyu Sun$^{1,2}$, Lei Qi$^{3}$, He Xu$^{4}$\\
\textit{\textbf{Affiliations}}\\
$^1$ Key Laboratory of Big Data \& Artificial Intelligence in Transportation (Ministry of Education), School of Computer and
Information Technology, Beijing Jiaotong University, Beijing, China\\
$^2$ Institute of Automation, Chinese Academy of Sciences, Beijing, China\\
$^3$ Image Technology Department, Transition, Shanghai, China\\
$^4$ College of Information Science and Engineering, Hohai University, Changzhou, China

~\\

\noindent \textit{\textbf{Team name:}} NTR\\
\textit{\textbf{Members:}}
Jiachen Tu$^1$ (jtu9@illinois.edu)\\
Guoyi Xu$^1$ (ericx3@illinois.edu)\\
Yaoxin Jiang$^1$ (yaoxinj2@illinois.edu)\\
Jiajia Liu$^1$ (ciciliu2@illinois.edu)\\
Yaokun Shi$^1$ (yaokuns2@illinois.edu)\\
\textit{\textbf{Affiliations}}\\
$^1$ University of Illinois at Urbana-Champaign, Champaign, IL 61820, USA
\section{Conclusion}

We introduced the NTIRE 2026 RAIM Track 2 challenge focused on multi-exposure fusion in dynamic scenes, emphasizing real-world HDR challenges like motion and misalignment. 
By combining a dynamic-scene dataset, a transparent evaluation protocol, and organizer-side reproducibility verification, we collect several methods that are not only accurate on the leaderboard but also robust in real-world use.
All participating teams contribute impressive methods and thoughtful insights.
Results show top submissions balanced fidelity and perceptual quality across varied conditions, highlighting the ongoing challenge of dynamic artifacts. 
We hope this challenge and the released resources will drive further research and the development of reliable, high-quality fusion methods for dynamic computational photography.

\noindent\textbf{Acknowledgement.} This work was partially supported by the Humboldt Foundation. We thank the NTIRE 2026 sponsors: OPPO, Kuaishou, and the University of Wurzburg (Computer Vision Lab).

{
    \small
    \bibliographystyle{ieeenat_fullname}
    \bibliography{main}
}

\end{document}